\documentclass[10pt,twocolumn,letterpaper]{article}

\usepackage{cvpr}
\usepackage{times}
\usepackage{epsfig}
\usepackage{graphicx}
\usepackage{amsmath}
\usepackage{amssymb}
\usepackage{booktabs}
\usepackage{multirow}
\usepackage{tabulary}

\usepackage{bigdelim}
\newcolumntype{P}[1]{>{\centering\arraybackslash}p{#1}}  

\usepackage{xr}
\usepackage[utf8]{inputenc}

\usepackage[pagebackref=true,breaklinks=true,letterpaper=true,colorlinks,bookmarks=false]{hyperref}

\cvprfinalcopy 

\def\cvprPaperID{****} 

\ifcvprfinal\fi
\begin{document}

\title{Contextual Classification Using Self-Supervised Auxiliary Models for Deep Neural Networks}

\author{Sebastian Palacio \quad Philipp Engler \quad Jörn Hees \quad Andreas Dengel \\
German Research Center for Artificial Intelligence (DFKI)\\
TU Kaiserslautern \\
\texttt{first.last@dfki.de}}

\maketitle


\begin{abstract}
Classification problems solved with deep neural networks (DNNs) typically rely on a closed world paradigm, and optimize over a single objective (\eg, minimization of the cross-entropy loss).
This setup dismisses all kinds of supporting signals that can be used to reinforce the existence or absence of a particular pattern.
The increasing need for models that are interpretable by design makes the inclusion of said contextual signals a crucial necessity.
To this end, we introduce the notion of Self-Supervised Autogenous Learning (SSAL) models.
A SSAL objective is realized through one or more additional targets that are derived from the original supervised classification task, following architectural principles found in multi-task learning.
SSAL branches impose low-level priors into the optimization process (\eg, grouping).
The ability of using SSAL branches during inference, allow models to converge faster, focusing on a richer set of class-relevant features.
We show that SSAL models consistently outperform the state-of-the-art while also providing structured predictions that are more interpretable.
\end{abstract}


\section{Introduction}
\label{sec:introduction}

Machine learning models tackling classification problems are isolated in nature \ie, they are defined, and operate under a closed world paradigm~\cite{bendale2015towards} where all possible inputs belong to one out of multiple but finite pre-defined classes.
This simplification goes against emerging needs for more interpretable models~\cite{doshi2017towards,rudin2019stop} potentially harming performance, as humans naturally rely on external, complementary knowledge to find corroborating or conflicting evidence for a particular decision.
Our brains process information in a non-linear fashion, aggregating heterogeneous stimuli that converge to a unified interpretation or action.
Closed world models are thereby semantically disconnected from the patterns we may deem reasonable, making the quest for explanations an ill-posed endeavor.
The field of adversarial perturbations is a good example of this semantic gap for explanability~\cite{szegedy2013intriguing}.
Despite having input samples that preserve all the perceptually relevant information, adversarially perturbed samples can be misclassified with high probability.
The effectiveness of adversarial attacks provides a strong body of evidence that patterns extracted by neural networks are effective but fundamentally different from the ones we are able to understand.

\begin{figure}[t]
\centering
    \includegraphics[width=\linewidth]{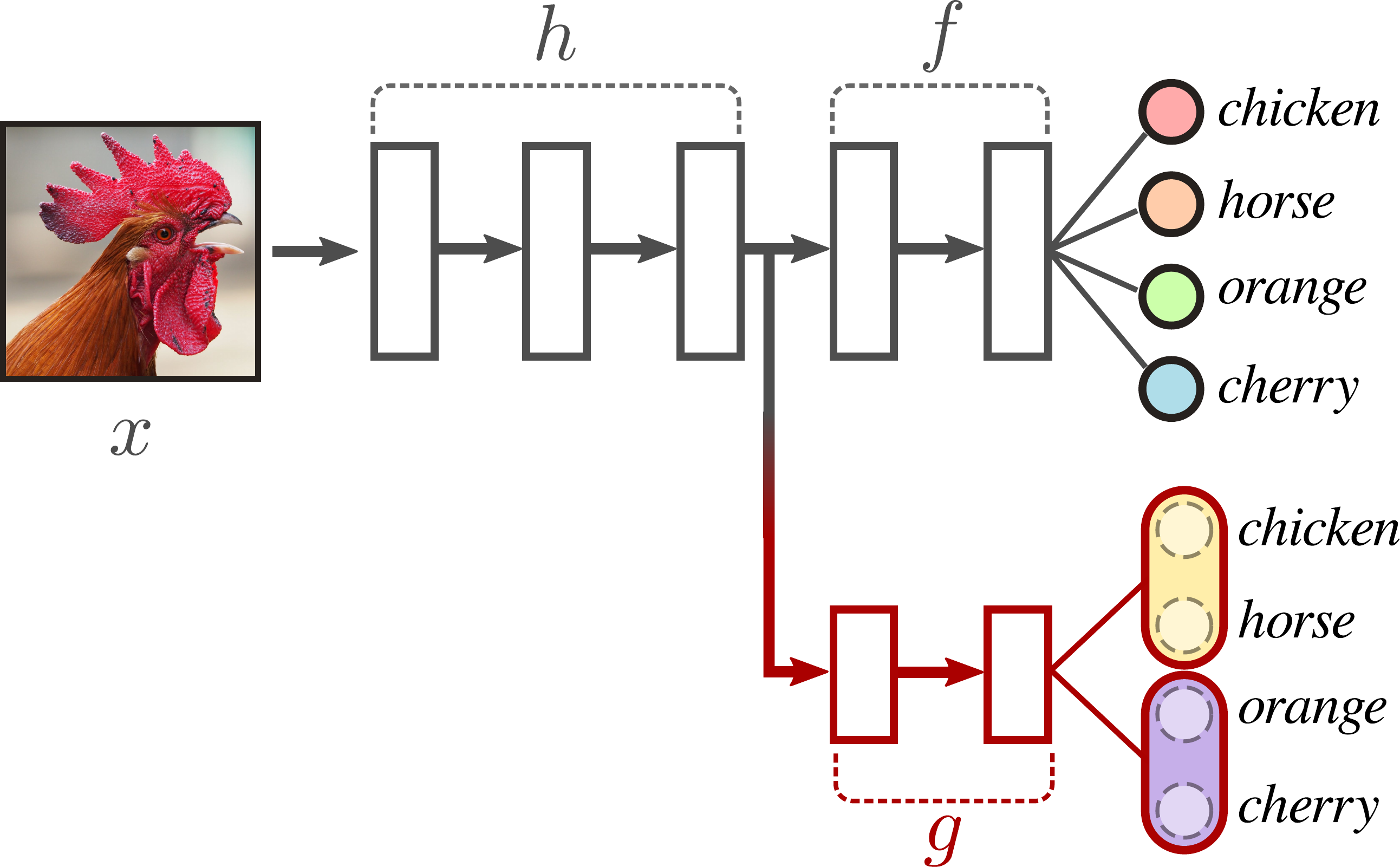}
	\caption{Overview of a SSAL model. Starting from a common feature representation $h$, a supervised goal $f$ and an auxiliary branch $g$ are used for training and prediction. Training objective for $g$ is derived from the original labels used for $f$ following a mutually exclusive grouping.}
	\label{fig:overview}
\end{figure}

Is there a way to embed context signals into the training process of a neural network without resorting to additional ground-truth?
Contextual information can be of course collected alongside class labels, but what exactly should that context be, is non-trivial and costly.
In this work, we propose the use of auxiliary classifiers to solve a surrogate objective that is still closely related to the original task.
Intuitively, we design the auxiliary task based on a simple characteristic of independent classification problems: if a model can classify a set of disjoint fine-grained classes, it should also be able to classify an arbitrary grouping of those classes.
The architecture of a traditional model can be hence modified by adding a symbiotic auxiliary classifier that shares a common feature representation, but optimizes the grouping objective instead (Figure~\ref{fig:overview}).
For prediction, a combination of both outputs is possible through an element-wise (Hadamard) product or via a learned linear combination.

We describe the auxiliary task as ``\emph{autogeonous}'' as it is self-supervised \ie it does not rely on additional annotations, and it is derived from a source within the dataset, namely the original labels.
We refer to the use of auxiliary classifiers using such a surrogate objective as \emph{Self-Supervised Autogenous Learning} (SSAL).
The close relationship between SSAL tasks and the main task allows the main model to benefit from auxiliary classifiers both during training and inference.

From the standpoint of set theory, it is easy to see how the main and auxiliary objectives are aligned while expressing fundamentally different goals.
Given a set of fixed class labels $\mathcal{Y} = \{y_1, y_2, \dots, y_k\}$, a labeled sample $(x_i, y_i)$, the singleton $\{y_i\} \subset \mathcal{Y}$, and the classification result $f(x_i) = y$, the prediction of $f$ is \emph{correct} iff $y \in \{y_i\}$.
A corollary of this classification setup is that the conditions for correctness remain unaltered for $f(x_i)$ when a second set $\mathcal{Y}_G = \{y_i\} \cup \{y_j\}$ with $y_i \neq y_j$ is considered.
SSAL corresponds to an auxiliary classifier $g$ that explicitly focuses on the relationship $g(x_i) = y$ is \emph{correct} iff $y \in \mathcal{Y}_G$.

The benefits of an SSAL paradigm is threefold: (1) it acts as regularizer for the original architecture, (2) the contextual nature of the surrogate objective provides supporting evidence that aligns with human expectations, therefore being interpretable by design.
Finally, (3) we show through a set of comprehensive experiments on CIFAR100~\cite{krizhevsky2009learning}, TinyImagenet~\cite{tinyimagenet}, and Imagenet~\cite{russakovsky2015imagenet}, that the joint training regime consistently yields superior accuracy even after controlling for model size.

\section{Related Work}
\label{sec:relatedwork}

\textbf{Early Work}: The notion of auxiliary classifiers for neural networks can be tracked down back to the early days of machine learning.  
In 1990, Abu \etal~\cite{abu1990learning} proposed the use of ``hints'' \ie additional knowledge about an objective for neural networks.
These hints were represented as an additional gradient term for a Multi-Layer Perceptron trained via back-propagation~\cite{rumelhart1983learning}.
They concluded that the use of hints could allow networks to converge faster as the set of potential solutions was constrained further.

\textbf{Multi-Task Learning}: This idea was later expanded to what is currently known as Multi-Task Learning (MTL)~\cite{caruana1997multitask}.
In this scenario, a network is trained jointly on multiple tasks that are not necessarily aligned but still share some commonalities, and therefore can benefit from a joint representation.
However, MTL falls short when defining metrics quantifying the similarity between tasks.
Furthermore, the result of MTL is a model that can solve two problems but the heterogeneity of the domains often precludes the possibility of a joint prediction.
Although recent advances on MTL have demonstrated that jointly learning disparate tasks does not harm performance~\cite{kaiser2017one}, measuring the net benefit for all tasks still depends on a perceived ad-hoc similarity between them.
The symbiotic interaction between the two objectives is thereby obscured, making it difficult to establish if one or both tasks benefit from one another (\ie if the relation is mutualistic, commensalistic or parasitic).
Some surprising and rather unexpected relationships between tasks have been reported in the literature, showing that the degree of relatedness between tasks is not trivial to assess.
For instance, Lee \etal~\cite{lee2017unsupervised} showed how a video classifier trained on finding the right order of a clip of shuffled frames was advantageous for fine-tuning on action recognition, image classification and object detection tasks.
Similarly, Vondrick \etal~\cite{vondrick2018tracking} also found that an effective object tracker can be obtained by training on a frame-colorization task.
In fact, a recent study of the relationships between a plethora of visual tasks explicitly highlights the non-triviality of the relationships between some of them~\cite{zamir2018taskonomy}.

\textbf{Hierarchical Priors}: One prominent use of auxiliary classifiers, beyond the scope of MTL, is representing hierarchical knowledge.
As the categories of classification problems are often semantically organized this way (\eg objects, places, animals, species, breeds), some work has focused on the benefits of encoding said priors.
The hierarchical relations can be known a-priori and used for adjusting a prediction by modeling relations of exclusion or subsumption~\cite{deng2014largescale}.
Alternatively, hierarchical relationships can be learned alongside the model to compensate for classes with a small number of samples~\cite{srivastava2013discriminative} or a posteriori where labels for a ``student'' model are represented by the ones learned previously by a ``teacher'' classifier~\cite{hinton2015distilling}.
Our approach focuses on relationships that are semantically present (through the original labels) but do not require additional annotations for the auxiliary task.
This implies that the relationships are innate (\ie, not learned), thereby not prone to limitations in the models or training schemes.

\textbf{Regularizing Branches}: Another well-known purpose of auxiliary classifiers has been the stabilization of gradient flow for very deep neural networks.
Most prominently, auxiliary classifiers were used for training different iterations of the Inception architecture~\cite{szegedy2016rethinking,szegedy2015going}.
Here, auxiliary networks were small parallel branches that used the same training objective as the main network.
These branches were only used during training and the reported benefits include faster convergence, more stable gradients and regularization.
More recently, models constructed via Neural Architecture Search also made use of auxiliary classifiers in a similar fashion~\cite{zoph2018learning}.
All of these examples use auxiliary classifiers exclusively for training, resort to the exact same loss function, and are not taken into account for prediction.

\textbf{Heterogeneous Surrogate Constraint}: Instead of one auxiliary branch with the same classification objective, benefits have been reported where small binary classifiers are attached at each layer~\cite{lee2015deeply}.
These binary branches optimize an objective that measures whether features at each layer are discriminative for the main prediction \ie, they yielded a true positive or a false negative prediction.
Alternatively, a reconstruction objective imposed to the original supervised cost (instead of the feature relevance score) was shown to improve classification as well~\cite{zhang2016augmenting}.
Our work is similar in that it also relies on a different auxiliary objective which requires no extra labels.
However, we use the notion of grouping which preserves more information from the original labels than the notion of feature relevance or input reconstruction.
This way, the alignment of the classification objective and the SSAL branch provides outputs that are explicitly and directly interpretable.

\textbf{Groups as Auxiliary Prior}: the idea of joining classes together has been exploited to improve upon a classification objective.
By grouping classes that fall under a more general semantic term (\eg, ``\textit{cat}'' and ``\textit{dog}'' are both ``\textit{animals}''), a data-augmentation scheme can easily mine additional data samples that relate to the term subsuming the included labels (the super-term) using a search engine~\cite{xie2015hyper}.
Both the crawled data and the original dataset are passed through a network with two corresponding branches and trained jointly.
Note that data for the ``auxiliary'' task (the branch for super-classes) is disjoint to the one used for the fine-grained task.
Moreover, there is no explicit correspondence between the classification of the super-class and the original class.
A different approach starts by assigning the original labels to visually similar groups and training a dedicated feature extractor for each one.
At the same time, a soft-gating mechanism is trained to decide which specialized feature extractors should be used, to finally combine their features into one prediction~\cite{mullapudi2018hydranets}.
In this case, there are no auxiliary objectives (grouping is a priori) and there is only one loss with a single prediction per sample.

Instead of a gating mechanism, HD-CNNs~\cite{yan2015hd} utilize a coarse classifier to control a set of specialized branches.
Due to the conditional re-routing of samples based on the coarse classifier or the soft-gating, training these models needs to be adaptive and multi-step, the risk of overfitting increases (specialized networks rely on fewer samples), and the computational cost goes up considerably as more fine-grained classifiers are used.

In contrast, we opt for a much simpler setup that is not affected by the number of coarse groups (in terms of computation), beyond the dimensions of the output layer.
Our proposed network can be trained jointly and end-to-end using standard optimization algorithms, with no conditional re-routing or special regularization mechanisms.


\section{Methods}
\label{sec:methods}

In this section, we describe the algorithmic components from SSAL and how they integrate into a traditional classification problem for training and prediction.
There are four main components to discuss: grouping criterion, architectural design, training objectives and joint prediction.
For each of these components, we introduce emergent meta-parameters that need to be considered during evaluation.

\subsection{Grouping Criterion}
\label{subsec:grouping}
We propose that the autogenous auxiliary objective be based on a grouping of the original classes.
Modeling groups explicitly allows a classifier to learn the property of subsumption; a proven useful mean to generate explanations in formal verification systems~\cite{mcguinness1995explaining}.
To this end, we use a clustering algorithm based on similar principles than the one used by Yan \etal~\cite{yan2015hd} but imposing a constrain that ensures balanced clusters.
Concretely, given a set of classes $\mathcal{Y}_c = \{y_1, y_2, \dots y_c\}$ we define $\mathcal{Y}_k$ as a partition of $\mathcal{Y}_c$ into $k$ subsets.
The grouping starts by constructing a distance matrix $\mathcal{D}_c$, based on the confusion matrix from a pre-trained model.
Given a normalized confusion matrix $F$ with the diagonal set to zero, the distance matrix $\mathcal{D}_c$ is constructed by subtracting 1 from it and then making it symmetric by averaging the off-diagonals (Equation~\ref{eq:distancematrix}).

\begin{align}
\hat{D} &= 1 - F
\label{eq:sim2dist}
\\
\mathcal{D}_c &= \frac{1}{2} (\hat{D} + \hat{D}^T)
\label{eq:distancematrix}
\end{align}

Each cluster is initialized with one of the $k$ labels with the highest average distance to all other labels.
The next label in $\mathcal{Y}_c$ to be assigned will be the one with the smallest distance to a cluster currently holding less than $c/k$ elements.
In case of a tied metric w.r.t.~a cluster, a random one among those is used for the assignment.
Note that the distance matrix $\mathcal{D}_c$ can be turned into a similarity measure by omitting the inversion of $F$ \ie, skipping Equation~\ref{eq:sim2dist}.

The output of this algorithm is a mapping $\gamma: \mathcal{Y}_c \rightarrow \mathcal{Y}_k$ assigning a single group label to each of the original ground-truth labels.
This way, each sample in a labeled dataset is modeled as a triplet $(x_i, y_i, \gamma(y_i))$ representing the input sample, the ground-truth label and the group label it has been assigned to respectively.
A more detailed description of the clustering algorithm, can be found in Section \ref{sup:clustering} of the supplementary material.

There are two meta-parameters that we consider for grouping, namely the number of groups to map to, and the criterion used for grouping.
While the former is expressed by an integer $2 \le k \le C/2$, the latter can prioritize either joining or splitting visually similar ground-truth labels (by controlling how $\hat{D}$ is computed).

\subsection{SSAL Architectural Design}
The proposed model follows the structure of a hard parameter sharing architecture for MTL with three main components, as shown in Figure~\ref{fig:overview}.
First, an initial, shared branch $h$ is in charge of extracting low-level features.
Next, these common features are fed into two branches $f$ and $g$ with different classification objectives: one with the original ground-truth classification objective, while the second branch optimizes over the group labels.
Given an input sample, the ensemble model will output a prediction for the original classification target $f(x)$ and a prediction for the auxiliary task based on grouping $g(x)$.

In practice, these architectures are realized by taking a traditional classifier like Resnet50~\cite{he2016deep}, and attaching an ancillary classifier (with a group objective) at some point in-between the layers of the original model.
The specific layer disposition for both auxiliary and original models depends on the experiment but in essence, a mixture of convolutional and pooling layers are used.
A more detailed specification of all networks used in this work can be found in Section~\ref{sup:networks} of the supplementary material.

Under this last perspective, an important meta-parameter of the architecture is the point at which the auxiliary classifier attaches to the original model.
Having a junction in earlier layers allows both branches to work with generic, lower-level features but leaves little room for those features to be regularized by the updates from both branches.
Another possible meta-parameter is the number of auxiliary classifiers that can be attached.
In that case, we refer to a set of (possibly different) groupings $\gamma_1, \gamma_2, \dots$ based on the ground-truth labels in $\mathcal{Y}_c$ for which a dedicated auxiliary branch $g_1, g_2, \dots$ is used.

\subsection{SSAL Training}
Training relies on traditional end-to-end backpropagation using mini-batch SGD.
Both branches $f$ and $g$ are trained jointly, and their individual errors are measured using cross-entropy.
Note that there is no unified prediction at this point and the losses for each branch are only added together to force an single update of the entire parameter space, including the common initial feature extractor $h$.
The sum is controlled by weights $\lambda_1$ and $\lambda_2$ as shown in Equation~\ref{eq:loss}:

\begin{equation}
\mathcal{L} = \lambda_1\, \mathcal{L}_f + \lambda_2 \mathcal{L}_g
\label{eq:loss}
\end{equation}

where $\mathcal{L}_f$ and $\mathcal{L}_g$ are the cross-entropy losses for $f$ and $g$ respectively.

\subsection{SSAL Prediction}
\label{subsec:prediction}
One of the main novelties of this work is the use of the auxiliary classifier for prediction.
To this end, we consider two alternatives for calculating a joint prediction.

\textbf{Joint Probability}: the final prediction is represented as the joint probability of the original prediction and the auxiliary classifier such that $P(y|x) = \textit{softmax}(f_i(x) \cdot g_{\gamma(i)}(x))$, where $f_i$ is the i-th output dimension of $f(x)$, $g_{\gamma(i)}(x)$ is the output dimension of the auxiliary branch associated with the original label at $i$ and $\cdot$ represents a scalar product.
When more than one auxiliary classifier is used, the output of all auxiliary branches $g_i$ is raised to a power $\eta \in (0, 1]$.

\textbf{Learned Linear Combination}: Predictions from $f(x)$ and $g(x)$ are concatenated and then used to train a linear classifier with the same number of outputs as there are labels in the original ground-truth.
Both $f$ and $g$ are assumed to be trained already, and the linear classifier is hence trained separately.

We also evaluate the prediction of $f(x)$ alone as a baseline.
This way, we can establish the influence that jointly training the auxiliary classifier has had in the performance of the branch with the original classification problem.
In other words, this baseline evaluation measures the inductive bias of the auxiliary classifier.


\section{Experiments}
\label{sec:experiments}

In this section we describe the datasets, meta-parameters, baselines and performance experiments to support and quantify the benefits of SSAL models.

%
%
\subsection{Datasets}
We conduct experiments on three different image classification datasets with varying degrees of complexity:

\textbf{CIFAR100}~\cite{krizhevsky2009learning}: extension of CIFAR10 where 60\,000 color images of size 32x32 belong to 100 different classes of fine-grained objects or animals.
The training and test set contain 50\,000 and 10\,000 images respectively.

\textbf{TinyImagenet}~\cite{tinyimagenet}: 110\,000 color images of size 64x64 split into 200 natural categories \eg, animals, food, furniture.
They are divided into 100\,000 samples for training and 10\,000 for validation.
The official testing set does not provide labels, hence we take a small portion of the training set for development and report results on the validation set.

\textbf{Imagenet}~\cite{russakovsky2015imagenet}: One of the largest image classification datasets available.
Image size is variable but samples are commonly downscaled to 300x300 pixels.
They comprise over 1.2M images across 1000 categories.
Similarly to TinyImagenet, we use the 50\,000 validation samples for testing and in turn, take a small portion of the training set for any validation that is required.

%
%
\subsection{SSAL Meta-Parameters}
As mentioned in Section~\ref{sec:methods}, SSAL models introduce a variety of meta-parameters requiring additional consideration.

\textbf{Layer Architecture}: as mentioned earlier, multiple architectures are used depending on the dataset and the objective of the experiment.
We base our evaluations and SSAL models on five high-performance architectures: Resnet18~\cite{he2016deep}, Resnet50~\cite{he2016deep}, Wide-Residual-Networks (WRNs)~\cite{zagoruyko2016wide}, Squeeze and Excitation Nets (SENets)~\cite{hu2018squeeze}, and DenseNets~\cite{huang2017densely}.
For things like the architecture of auxiliary branches, we use a combination of blocks comprising convolutional, pooling, batch-normalization and inception-like layers.
When a meta-parameter search on these architectural elements is required, we use a small portion of the training set for validation, before evaluating on the corresponding test set.

\textbf{Grouping Criterion and Number of Groups}: we train a SSAL model based on Resnet18 for CIFAR100 and TinyImagenet.
The auxiliary classifier consists of four convolutional layers with batch-normalization and ReLU activation, a global average pooling, two fully connected layers and a final linear combination with softmax normalization.
The size and number of convolutional filters, and the number of fully connected neurons were determined via meta-parameter search.
See Section~\ref{sup:auxiliaryarchs} of the supplementary material for further details about said parameters.

We use a single SSAL branch (\ie an auxiliary classifier) with either 2, 4, 10 or 20 groups, and a grouping criterion that either splits or joins visually similar classes following the computation outlined in Section~\ref{subsec:grouping}.
The model prediction is done by computing the joint probability as proposed in Section~\ref{subsec:prediction}.
The auxiliary classifier attaches to the main network after the first max-pooling, and before the first residual block.

\begin{figure}[t]
\centering
    \includegraphics[width=\linewidth]{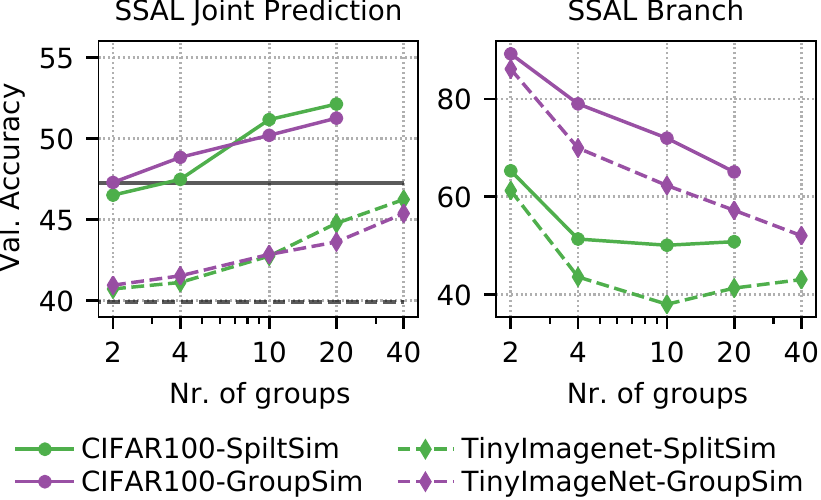}
	\caption{Accuracy when varying the number of groups and grouping criterion on CIFAR100. Left: joint prediction of the SSAL model. Right: classification of the auxiliary classifier alone  .}
	\label{fig:cifar100resnet10groups}
\end{figure}

Results in Figure~\ref{fig:cifar100resnet10groups} show a constant improvement of the combined classification error as the number of groups increases.
Although grouping visually similar classes yields an initial small advantage compared to the ``splitting'' criterion, this tendency inverts when the number of groups reaches 10 and 20.
The pattern, albeit some marginal fluctuations, is preserved for both CIFAR100 and TinyImagenet.

\begin{figure}
\centering
    \includegraphics[width=\linewidth]{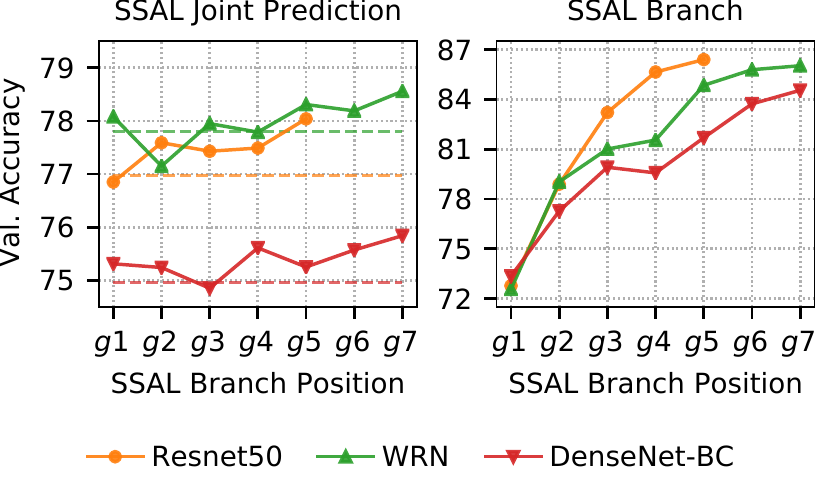}
	\caption{Classification accuracy when the position of the auxiliary classifier varies w.r.t.~the main network. Left: joint prediction of the SSAL ensemble. Right: classification of the auxiliary classifier (trained on 20 visually similar groups).}
	\label{fig:cifar100postion}
\end{figure}

This first experiment suggests that having more groups is beneficial and that either splitting or joining visually similar classes contribute to a better performance at a similar rate.

\begin{figure}[t]
\centering
    \includegraphics[width=\linewidth]{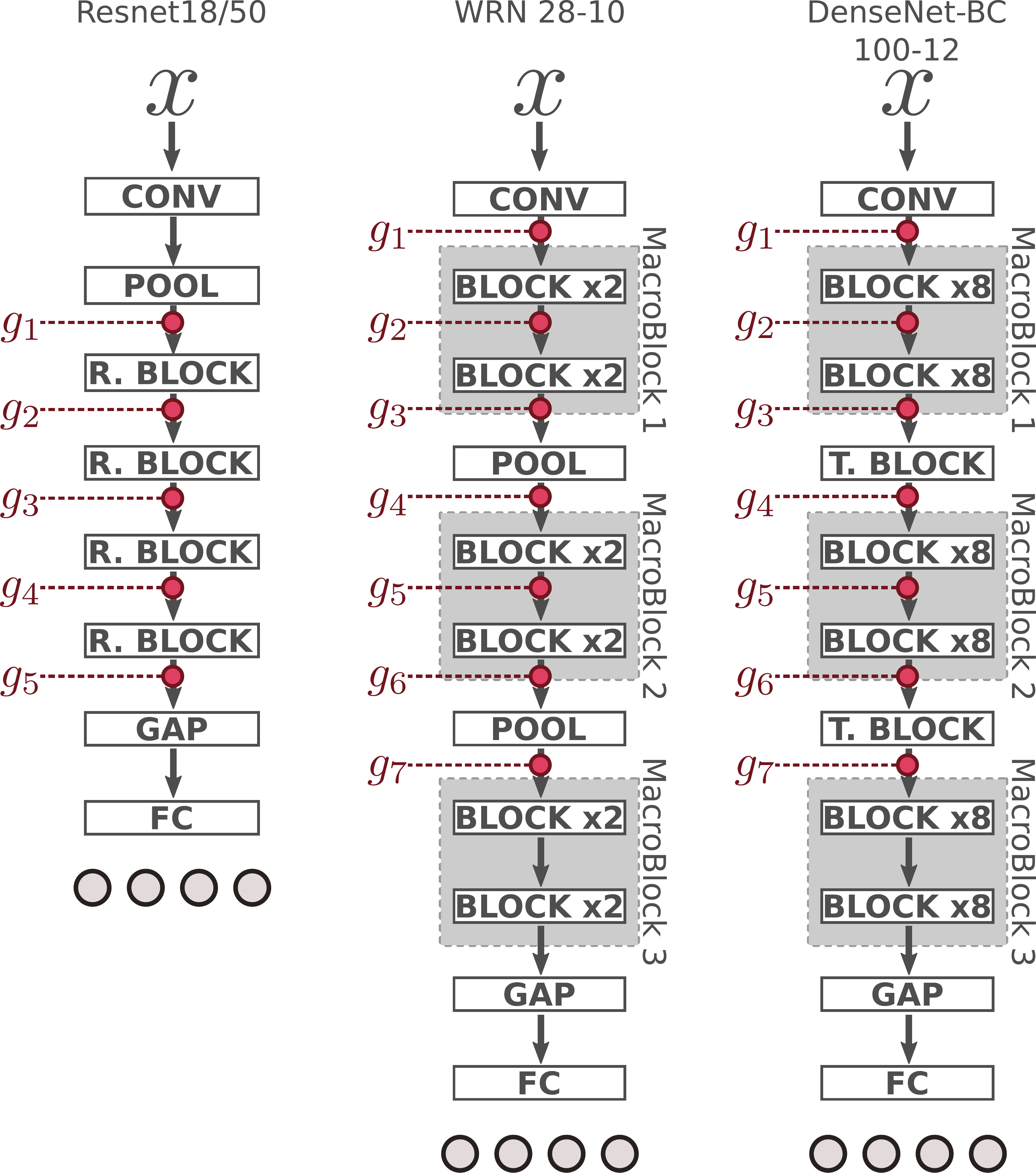}
	\caption{Positions where SSAL branches are being attached to the main network (red circles). Only one auxiliary branch is evaluated at a time.}
	\label{fig:attachmentpoints}
\end{figure}

\textbf{Position of the Auxiliary Classifier}: we use a similar setup based on Resnet50 for CIFAR100 and vary the point at which the auxiliary classifier is attached.
Each of the four residual blocks in the original network is considered an atomic unit.
We evaluate the effects of attaching the auxiliary classifier after each one of said blocks.
The architecture of the auxiliary classifiers remain the same, except for the number of channels in the first layer which increases as the point of attachment lies deeper in the original network.
They all optimize over the same 20 groups joining visually similar classes, and the final prediction is done via the joint probability.
The same experiment is conducted using two different classifiers: a Wide Residual Network (WRN 28-10) and a DenseNet (DenseNet-BC 100-12).
Seven points of attachment at different depths are selected for each network.
These points include paths that lie before, after and in-between macro-blocks (see Figure~\ref{fig:attachmentpoints}).

Results in Figure~\ref{fig:cifar100postion} show that the position of the auxiliary classifier w.r.t.~the main model has a tendency to perform best when the auxiliary classifier is attached at deeper layers of the original network.
This behaviour corresponds directly with the performance of the SSAL branch itself, which shows higher performance when it has been attached at a deeper stage within the architecture.


\textbf{Number of Auxiliary Classifiers}: to evaluate the influence of attaching more than one auxiliary classifier to the main model, we train Resnet18 on CIFAR100 while either one or two auxiliary classifiers $g_1, g_2$ are attached.
The auxiliary classifiers are both composed of two convolutional layers with batch-normalization and ReLU followed by an inception-like layer, global average pooling and a linear output layer with softmax normalization.
The auxiliary branch $g_1$ is placed after the first residual block and optimizes over 20 visually similar groups.
The $g_2$ counterpart is placed after the second residual block and optimizes over 50 visually similar groups.
Final prediction is based on the joint probability and an equal normalization power $\eta = 1$ is used.

Two similar experiments are conducted using Resnet50 and WRN.
For these two variants, three auxiliary branches are simultaneously attached.
Grouping is based on visual similarity and they optimize over an increasing number of groups: 20, 30 and 50 groups.
The normalization power is applied to all branches.
Results are summarized in Table~\ref{tab:cifar100number}.

\begin{table}[t]
\centering
\begin{tabular}{@{}lcccr@{}}
\toprule
          & $||g_i||$ & $\eta$ & Test Acc. (\%) & Parameters \\ \midrule
Resnet18  & 0         & -      & 75.67              & 11.23M     \\
          & 1         & 1.0    & 76.62             & 11.92M     \\
          & 2         & 1.0    & \textbf{78.23}     & 12.83M     \\ \midrule
Resnet50  & 0         & -      & 79.13              & 23.77M     \\
          & 1         & 1.0    & 79.70              & 25.07M     \\
          & 3         & 1.0    & 80.36              & 28.89M     \\
          & 3         & 0.3    & \textbf{80.69}     & 28.89M     \\ \midrule
WRN 28-10 & 0         & -      & 80.19              & 36.56M     \\
          & 1         & 1.0    & 80.96              & 38.19M     \\
          & 3         & 1.0    & 80.68              & 43.25M     \\
          & 3         & 0.4    & \textbf{81.08}     & 43.25M     \\ \bottomrule
\end{tabular}
\caption{Classification accuracy for models with $||g_i||$ SSAL branches on CIFAR100. Using more branches, together with regularization, improves performance. However the computational footprint also increases.}
\label{tab:cifar100number}
\end{table}

Increasing the number of SSAL branches does have a positive impact on performance, as long as the normalization power $\eta$ decreases when the number of SSAL branches increases.
Intuitively, the role of SSAL branches is one of verification and support rather than a predominant signal, and thereby outputs from this branches should be weighted down in scenarios when there are branches outnumbering the original classification network.
For Resnet18, adding two SSAL branches yields an accuracy of 78.2\%, a 2.6 pp.~improvement over the baseline.

%
%
\subsection{Alternative Baselines}
\label{subsec:baselines}
The use of auxiliary branches inevitably adds more raw capacity to the overall network by virtue of the extra trainable parameters.
We test whether the consistent boost in performance can be explained by the additional weights (Occam's razor) or if the introduction of the SSAL objective has merit on its own.

To this end, we train modified versions of Resnet18 on TinyImagenet that add more weights in various ways, matching or surpassing the number of parameters of a SSAL model.
We also compare models with the same architectural layout SSAL models have but training without the SSAL objective.

\textbf{WideResnet18}: has 50\% more filters across all convolutional layers.

\textbf{DeepResnet18}: adds four convolutional layers of 256 filters each with batch-normalization and ReLU activations before the first residual block.

\textbf{DWResnet18}: similar to DeepResnet18 but doubling the number of filters of the additional convolutional layers.

\textbf{GapCatNoSSAL}: based on a SSAL model but without the SSAL loss.
The output of the SSAL branch is concatenated to the GAP activation of the main classifier.

\textbf{CatFCNoSSAL}: based on the SSAL architecture but without the SSAL objective.
The output of the SSAL branch and the original network are concatenated and passed through a fully-connected layer with 2048 neurons.
This result is in turn passed through a linear combination for the final prediction.

\textbf{LinearComb}: this is a fully trained SSAL model, but instead of issuing predictions through a joint probability, both the auxiliary output and the prediction from the original classifier are concatenated together and used to train a separate linear classifier.

\textbf{SSAL}: classifier ensemble proposed in this work.
For the variant with one auxiliary classifier, the SSAL branch is placed after the first residual block ($g2$ in Figure~\ref{fig:attachmentpoints}) while the model with three SSAL branches correspond to attachment points for $g1, g2,$ and $g3$.

Networks are trained for 20 epochs with a triangular learning rate peaking at epoch 8.
Each experiment is repeated three times to account for initialization effects.
Results are summarized in Table~\ref{tab:baselines}.

\begin{table}[t]
\centering
\begin{tabular}{@{}lrrr@{}}
\toprule
                 & Val. Acc. (\%)         & Diff (pp.)     & Parameters        \\ \midrule
Resnet18         & 39.9 $\pm$ 0.3          & 0.0           & 11.2M             \\
WideResnet18     & 42.3 $\pm$ 0.3          & 2.4           & 25.3M             \\
DeepResnet18     & 43.1 $\pm$ 0.4          & 3.2           & 13.3M             \\
DWResnet18       & 43.7 $\pm$ 0.1          & 3.8           & 19.0M             \\ \midrule
GapCatNoSSAL     & 40.2 $\pm$ 0.3          & 0.3           & 15.6M             \\
CatFCNoSSAL      & 35.3 $\pm$ 0.8          & -4.6          & 13.6M             \\ \midrule
\textbf{LinearComb}  & \textbf{44.1 $\pm$ 0.1} & \textbf{4.2}  & 12.8M             \\
\textbf{SSAL x1} & \textbf{45.8 $\pm$ 0.2} & \textbf{5.9}  & 12.6M             \\
\textbf{SSAL x3} & \textbf{50.0 $\pm$ 0.4} & \textbf{10.1} & 15.6M             \\ \bottomrule
\end{tabular}
\caption{Baselines for SSAL models on TinyImagenet. Models that have a deeper architecture, wider layers, or lack the SSAL objective fail to reach the level of accuracy of SSAL models.}
\label{tab:baselines}
\end{table}

It is clear that adding more capacity to Resnet18 improves accuracy.
Capacity in the form of deeper layers shows better results than using wider layers, and a combination of both yields an overall improvement of up to 3.8 percentage points.
Using the same architectural disposition of a SSAL model but without the SSAL objective (*NoSSAL), worsen performance w.r.t.~the baseline, discarding this setup as the reason for improvements.
Overall, the use of SSAL objectives remains the most effective use of the extra weights and layers with an improvement over the baseline of 5.9 to 12.7 percentage points (8.9 points better than the best baseline, while keeping a lower paramater count).

\begin{figure}[t]
\centering
    \includegraphics[width=\linewidth]{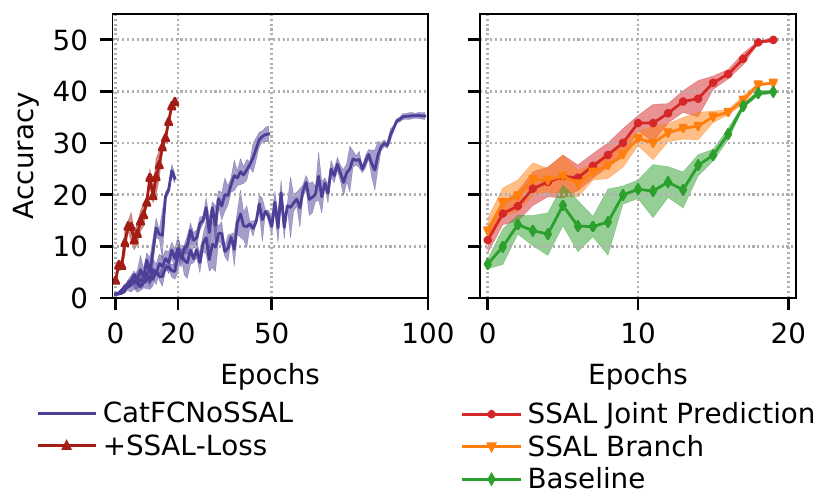}
	\caption{Classification accuracy of a SSAL Resnet18 (red) is not only higher, but its best performance is reached after training for fewer epochs than any baseline (purple).}
	\label{fig:convergence}
\end{figure}

\textbf{Training Convergence}: we measure the rate of convergence when training a SSAL model on CIFAR100, verifying that the SSAL objective has an aligned inductive bias which is not only beneficial for classification but it also requires less training steps.
We train the \emph{CatFCNoSSAL} baseline for 20, 50 and 100 epochs, and compare it with a fourth identical run, except that the SSAL objective is added to the training procedure.
Results in Figure~\ref{fig:convergence} (left) show the validation accuracy of these four systems.
Here we see that training with the SSAL objective drastically accelerates convergence.
Even after 100 epochs, an identical architecture is still unable to match the performance of its SSAL counterpart.
The accelerated convergence rate is also evident when comparing the training curves of the joint prediction of a SSAL-based pipeline against a baseline implementation with no auxiliary branches (Figure~\ref{fig:convergence} right).

%
%
\subsection{Improving Classification}
\label{sec:classification}
Based on the analysis of meta-parameters for SSAL models, we show that high accuracy is consistently attainable across a variety of well-known, thoroughly optimized architectures.

\textbf{CIFAR100}: We train SSAL models based on Resnet50, WRN, SENet and DenseNet on CIFAR100.
For each of these original architectures, we attach three SSAL branches with visually similar groups of 20, 33, and 50 groups.
To guarantee uniformity on the evaluation conditions, we have re-implemented all models and trained them from scratch so that the only difference between the original performance and the SSAL variant is the proposed surrogate objective.
Moreover, we report baselines from the original source (org), our own re-implementation (ours), and the \textit{LinearComb} setup from Section~\ref{subsec:baselines} (+LC).
For SSAL models, we report the accuracy of the original classifier \ie, using the SSAL branch during training but not for prediction (+TR), and the full SSAL prediction using the joint probability (+JP).
For further details about the architecture of the SSAL branches and the training setup, please refer to Section~\ref{sup:auxiliaryarchs} in the supplementary material.
Table~\ref{tab:cifar100accuracy} summarizes the results.

\begin{table}[t]
\centering
\begin{tabulary}{\textwidth}{@{}lcp{11pt}p{11pt}p{11pt}p{11pt}p{11pt}p{11pt}@{}}
\toprule
                     & \multicolumn{5}{c}{Val. Accuracy}                                                                                                  & \multicolumn{2}{c}{Params~(M)}                     \\ \midrule
\multicolumn{1}{p{11pt}}{} & \multicolumn{1}{p{11pt}}{org}    & \multicolumn{1}{p{11pt}}{ours} & \multicolumn{1}{p{11pt}}{+TR} & \multicolumn{1}{p{11pt}}{+JP} & \multicolumn{1}{p{11pt}}{+LC} & \multicolumn{1}{p{11pt}}{org} & \multicolumn{1}{p{11pt}}{SSAL} \\ \midrule
Resnet50             & -                               & 78.9                     & 79.7                    & 80.6                    & 80.2                    & 23.8                   & 28.9                    \\ 
SE-WRN~16-8          & \multicolumn{1}{p{11pt}}{80.9}  & 79.0                     & 79.0                    & 80.2                    & 80.0                    & 11.1                   & 14.9                    \\ 
WRN~28-10            & \multicolumn{1}{p{11pt}}{80.8}  & 80.1                     & 80.6                    & \textbf{81.0}           & 80.7                    & 36.6                   & 38.2                    \\ 
DenseNet~190-40      & \multicolumn{1}{p{11pt}}{82.8}  & 81.1                     & 81.8                    & \textbf{83.2}           & 83.1                    & 26.1                   & 38.3                    \\ \bottomrule
\end{tabulary}
\caption{Classification accuracy for multiple high-performance architectures on CIFAR100. Adding the SSAL objective consistently yields higher performance.}
\label{tab:cifar100accuracy}
\end{table}

These experiments show that training with the auxiliary classifier consistently yields better performance.
The inductive bias of the SSAL branch guides the classifier even when the auxiliary output is not used for prediction.
Performance improves even further when SSAL models issue a joint prediction.
Note that for WRN and DenseNet, the SSAL version outperforms the state-of-the-art that was originally reported, notwithstanding the weaker baseline it starts from.

\textbf{Imagenet}:
To test the effects of SSAL branches on large scale problems, we train a Resnet50 on Imagenet (ours), and compare it with a corresponding SSAL model with three auxiliary branches.
As in the previous experiment, they use visually similar classes with 200, 334, and  500 groups, and report values for training with SSAL only (+TR), joint prediction (+JP) and using the \emph{LinearComb} setup from~\ref{subsec:baselines} (+LC). 
We also evaluate on a \emph{GapCatNoSSAL} baseline (GC) from~\ref{subsec:baselines} which has a similar architecture but no SSAL objective.
Table~\ref{tab:imagenetacc} shows how, once again, a SSAL model is able to outperform the original baseline by almost 1.5 p.p..
Table~\ref{tab:sota}, compares our results with recently proposed state-of-the-art classifiers that convey contextual information in the loss function, use other kind of auxiliary classifiers or rely on different hierarchical priors for training.

\begin{table}[t]
\centering
\begin{tabulary}{\textwidth}{@{}p{25pt}p{30pt}p{30pt}p{30pt}p{30pt}p{30pt}@{}}
\toprule
                  & ours (org)        & +TR               & +JP                         & +LC               & GC             \\ \midrule
Top-1             & 75.5   & 76.4  & \textbf{76.9}   & 76.7  & 75.7 \\
Top-5             & 92.7   & 93.3  & \textbf{93.7}   & 93.4  & 92.7 \\ \bottomrule
\end{tabulary}
\caption{Accuracy for SSAL Resnet50 on Imagenet. Experiments are run 3 times. Standard deviation is 0.1.}
\label{tab:imagenetacc}
\end{table}

\begin{table}[t]
\centering
\begin{tabulary}{\textwidth}{@{}p{85pt}p{55pt}p{55pt}p{55pt}@{}}
\toprule
                                        & CIFAR100        & Imagenet                                   \\
\midrule
HD-CNN~\cite{yan2015hd}                 & 65.64           & 68.66 (-)*                                 \\
HydraNets~\cite{mullapudi2018hydranets} & 76.25           & 73.20 (-)*                                 \\
COT~\cite{chen2018complement}           & 79.46           & 75.60 (-)                                  \\
DSL~\cite{lee2015deeply,li2020dynamic}  & 81.95           & 76.12 (92.93)                              \\
DHM~\cite{li2020dynamic}                & 82.80           & 76.57 (93.24)                              \\
Aux.~Train~\cite{zhang2020auxiliary}    & 80.84           & 74.14 (-)*                                 \\
\textbf{SSAL (ours)}                    & \textbf{83.24}  & \textbf{77.00}  (\textbf{93.80})   \\
\bottomrule
\end{tabulary}
\caption{Top-1 accuracy of related state-of-the-art and SSAL models. Results for Imagenet are based on Resnet50 except the ones marked with *. Top-5 shown in parenthesis, if available.}
\label{tab:sota}
\end{table}

%
%
\subsection{Contextual Validation}
We show that predictions of SSAL models are more interpretable than regular DNNs thanks to the grouping objective of their auxiliary branches.
The use of heatmaps has been controversial as a mean to interpret a model's output because it can only point to the area of importance while leaving out information about the underlying features that elicit a high response~\cite{rudin2019stop}.
Labels within each SSAL group can be used to identify which low-level features are responsible for the prediction.

\begin{figure}[t]
\centering
    \includegraphics[width=\linewidth]{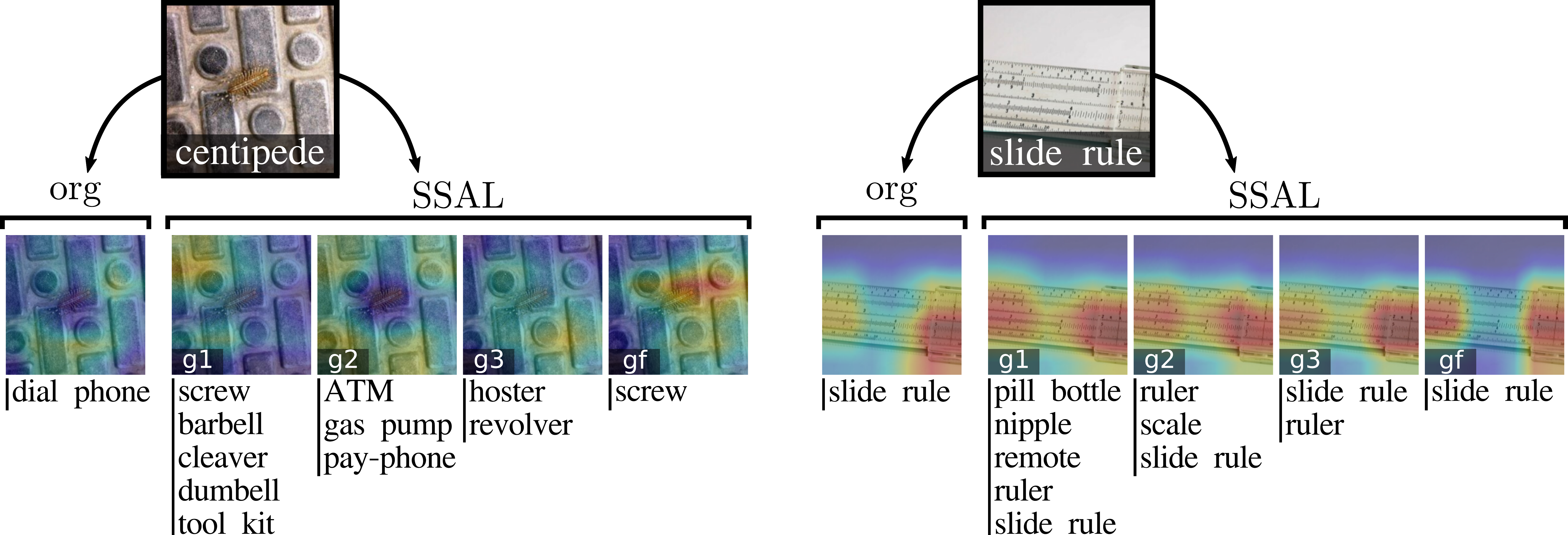}
	\caption{CAM \wrt~each auxiliary branch $g_i$ of the SSAL model. $g_f$ denotes the final classification output of the SSAL model, and \textbf{org} is the CAM of a normal Resnet50. Labels within predicted SSAL groups are shown below each branch.}
	\label{fig:ssal-cam}
\end{figure}

Figure~\ref{fig:ssal-cam} shows the Class Activation Mapping~\cite{zhou2016cnnlocalization} of two examples: a false-, and a true-positive.
For the former, predicted SSAL groups contain labels with metallic parts, and box-like shapes which correspond to areas with a strong activation.
For the slide rule, class labels in auxiliary groups like ``pill bottle'' or ``nipple'' (mouthpiece of a baby bottle) often depict the uniform markings found in rulers; a strong indication that these are precisely the salient features that guided this particular prediction.

In contrast, regular classifiers provide less nuanced insights where multiple interpretations are possible.
Examples in Figure~\ref{fig:ssal-cam} leave ample room for interpretation when predicting ``dial phone'' or even the true positive for ``slide rule'' (\texttt{org}).
More examples in the supplementary material.


\section{Conclusions}
In this work, we have introduced SSAL: a methodology for extending neural network architectures with auxiliary objectives that are related to the original task.
These objectives express low-level priors (\eg grouping the labels), do not require additional annotations, but derive from a pre-existing annotated set \ie, they are \textit{autogenous}.
SSAL models follow the structure of multi-task learning algorithms, therefore making a joint prediction possible based on the outputs from all branches in the model.
We show that the use of SSAL objectives consistently yields higher classification performance across several state-of-the-art classifiers like Resnets, DenseNets, SENets and WRNs for different datasets like CIFAR100, TinyImagenet and Imagenet.
The usefulness of the SSAL objective is validated through a comparison with several baselines including networks with similar architectural structure but no SSAL objective and networks with a comparable number of parameters.
Finally, we show how SSAL models leverage existing interpretability methods (\eg CAM) via the the low-level prior it was trained on, the model itself interpretable by design.

\paragraph{Acknoledgments:} This work was supported by the BMBF project ExplAINN (01IS19074), DeFuseNN (Grant 01IW17002) and the NVIDIA AI Lab program.

{\small
\bibliographystyle{ieee_fullname}
\bibliography{egbib}
}

\end{document}